\documentclass[final]{cvpr}

\usepackage{times}
\usepackage{epsfig}
\usepackage{graphicx}
\usepackage{amsmath}
\usepackage{amssymb}
\usepackage{subcaption}

\usepackage{booktabs}
\usepackage{tabularx}
\usepackage{colortbl}
\usepackage{multirow}
\usepackage{comment}
\usepackage{xfrac}
\usepackage{hyphenat}

\newcommand{\degree}[1]{${#1}^o$}
\def\360{\degree{360}}

\usepackage{calrsfs}
\DeclareMathAlphabet{\pazocal}{OMS}{zplm}{m}{n}
\newcommand{\layout}{\pazocal{S}}

\newcommand{\im}{\pazocal{I}}    
  
\newcommand{\obj}{\pazocal{M}}  
\newcommand{\loss}{\pazocal{L}}   
\newcommand{\Attention}{\pazocal{A}}
\newcommand{\vgg}{\Phi}

\newcommand*{\affaddr}[1]{#1} 

\newcommand*{\email}[1]{\small{\texttt{#1}}}

\usepackage[pagebackref=true,breaklinks=true,colorlinks,bookmarks=false]{hyperref}

\def\cvprPaperID{11} %
\def\confYear{CVPR 2021}
\usepackage{nopageno}

\begin{document}

\title{\vspace{-1em}PanoDR: Spherical Panorama Diminished Reality for Indoor Scenes}

\author{Vasileios Gkitsas \,\,  Vladimiros Sterzentsenko  \,\,  Nikolaos Zioulis  \,\, Georgios Albanis  \,\, Dimitrios Zarpalas \\
\affaddr{Centre for Research and Technology Hellas, Thessaloniki, Greece}\\
\email{\{gkitsasv,vladster,nzioulis,galbanis,zarpalas\}@iti.gr}
}

\twocolumn[{%
\renewcommand\twocolumn[1][]{#1}%
\maketitle

\begin{center}
    \centering    
    \includegraphics[width=\textwidth]{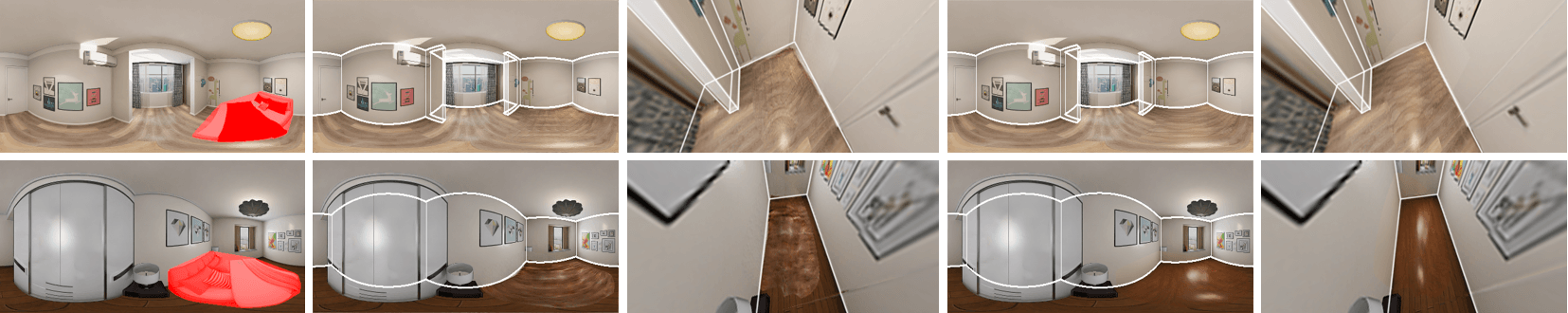}
    \captionof{figure}{Diminishing the highlighted (\textcolor{red}{red} mask) object in indoor spherical panorama images.
    White lines annotate the scene's layout in panorama and perspective views.
    Left to right: \textbf{i)} masked object to remove, \textbf{ii)} pure inpainting result of state-of-the-art methods (top row: RFR \cite{li2020recurrent}, bottom row: PICNet \cite{zheng2019pluralistic}), \textbf{iii)} perspective view of inpainted region by these methods better shows that they do not necessarily respect the scene's structural layout, \textbf{iv)} our panorama inpainting that takes a step towards preserving the structural reality, \textbf{v)} perspective view of inpainted region by our model, showing superior results both in texture generation and layout preservation.
    The results in this figure depict cases where RFR and PICNet provide reasonable structural coherence, and aim at showcasing our model's finer-grained accuracy.
    Figure~\ref{fig:qualitative} presents more qualitative examples where the structural in-coherency of RFR and PICNet is more evident.
    }
    \label{fig:teaser}
\end{center}%
}]

\begin{abstract}
   The rising availability of commercial \360 cameras that democratize indoor scanning, has increased the interest for novel applications, such as interior space re-design. 
   Diminished Reality (DR) fulfills the requirement of such applications, to remove existing objects in the scene, essentially translating this to a counterfactual inpainting task. 
   While recent advances in data-driven inpainting have shown significant progress in generating realistic samples, they are not constrained to produce results with reality mapped structures.
   To preserve the `reality' in indoor (re-)planning applications, the scene's structure preservation is crucial.
   To ensure structure-aware counterfactual inpainting, we propose a model that initially predicts the structure of a indoor scene and then uses it to guide the reconstruction of an empty -- background only -- representation of the same scene.
   We train and compare against other state-of-the-art methods on a version of the Structured3D dataset \cite{Structured3D} modified for DR, showing superior results in both quantitative metrics and qualitative results, but more interestingly, our approach exhibits a much faster convergence rate. 
   Code and models are available at \href{https://vcl3d.github.io/PanoDR/}{vcl3d.github.io/PanoDR/}.
\end{abstract}

\section{Introduction}
The advances in omnidirectional imaging are increasing their adoption in various sectors.
The introduction of consumer-grade spherical cameras, or mobile applications stitching moving camera videos into spherical panoramas, can enable new user experiences driven by the holistic capture capacity of \360 cameras.
One growing field driven by this holistic imaging property is indoor scene understanding as prominently demonstrated in the pioneering work of PanoContext \cite{zhang2014panocontext}.

PanoContext used a spherical panorama to estimate a room's layout, and has been recently succeeded by a large body of data-driven methods \cite{yang2019dula,sun2019horizonnet,zou2018layoutnet,zou20193d} to infer generalized indoor scene layouts from a single monocular \360 image. 
Indoor layouts represent a coarse geometry representation which can nonetheless be used to reconstruct entire buildings from overlapping captures \cite{pintore2016omnidirectional,pintore20183d,pintore2018recovering}.
In addition, the recent availability of real-world \cite{zioulis2018omnidepth,zioulis2019spherical,karakottas2019360} or synthetic \cite{jin2020geometric,Structured3D} spherical depth datasets  has enabled finer-grained geometry estimation from panoramas.

Apart from the structural understanding of indoor scenes, \360 images also offer a holistic contextual representation of a scene.
Be it either for the detection \cite{guerrero2020s} or segmentation \cite{zhang2019orientation} of objects, the combination of contextual and structural understanding can drive next-generation applications offering new forms of interaction.
One such application is the Augmented Reality (AR) aided planning of interior spaces, in the context of refurnishing, redecorating or retail, which relies on the simultaneous availability of both structural (layout, geometry) and contextual (object related) information.
However, an important shortcoming of AR in such a setting is its limitation to only enhance the real world with virtual elements via visual overlay.
Yet in the context of interior (re-)design, the removal of objects is also very relevant, which can be achieved through Diminished Reality (DR).
The latter can act either in a complementary manner to AR, allowing it to ``replace'' objects by first removing (DR), and then overlaying (AR), or by purely erasing them from the scene.

DR is an important technology which has been overlooked for \360 content.
It is very closely related to image and/or video inpainting which seeks to reconstruct missing image regions and has achieved impressive results using modern data-driven models.
Nevertheless, it also takes a step beyond traditional inpainting as it needs to respect the surrounding context in stricter ways, moving towards realistic reconstructions instead of plausible ones.
Indeed, image inpainting state-of-the-art currently focuses solely on photorealism \cite{Nazeri_2019_ICCV}, and even the richness of the plausible inpaints \cite{zheng2019pluralistic}, foregoing constraints about the alignment of the hallucinated content with the actual scene.

In this work, we focus on \360 DR in the context of interior (re-)design applications, relying on image inpainting and recently available datasets to reconstruct foreground occluded areas faithfully.
The contextual realism required to move beyond plausible inpaints in this case is reliant on the scene's structure, a very important cue for indoor space replanning and/or refurnishing that needs to be respected.
In summary our contributions are the following:
\vspace{-0.15cm}
\begin{itemize}
    \item We present the first, to the authors' knowledge, method for DR-oriented inpainting in \360 images using recently available panorama datasets with paired full and empty scenes (Figure~\ref{fig:dataset}).
    \vspace{-0.15cm}
    \item To preserve the structural reality when diminishing scenes we bridge image-to-image translation with generative inpainting, conditioning the inpainting results on the underlying scene structure.
    \vspace{-0.15cm}
    \item A DR model with fast convergence rate, significantly faster than state-of-the-art inpainting techniques, outperforming them in both photorealism and structural coherency.
\end{itemize}

\section{Related Work}

\textbf{Data-driven Inpainting.}
Context Encoders \cite{pathak2016context}, one of the first data-driven methods designed for image inpainting, combined an autoencoder with an adversarial loss in order to generate sharper images. 
Similarly, Iizuka et al.~\cite{iizuka2017globally} used two discriminators, one global and one local, to enforce photo-consistency, while also increasing the model's receptive field using dilated convolutions.
These early methods conditioned their predictions on both valid and masked inputs, leading to visual artifacts such as color discrepancy and blurriness.
To overcome such limitations, Liu et al.~\cite{liu2018image} introduced partial convolutions for image inpainting to prevent the accumulation of zeros in the encoded representations.
Yu et al.~\cite{yu2019free} extended this idea by proposing gated convolutions to learn the mask automatically which, combined with SN-PatchGAN \cite{miyato2018spectral, li2016precomputed}, achieved higher quality inpainting results.

Zeng et al.~\cite{zeng2019learning} showed that a pyramid-context encoder exploiting the information of different scales, improved the image completion result. 
The pyramid-context encoder progressively learned region affinity by attention from a high-level semantic feature map, transferring the learned attention to the low-level feature maps.
Another work relying on multi-scale feature fusion was the mutual encoder-decoder work of Liu et al.~\cite{liu2020rethinking}. 
CNN features from shallow and deep layers were used to represent the textures and structures of the input image, respectively.
Splitting these semantically different but complementary representations into two branches and jointly exploiting them to inpaint in multiple scales produces high quality results.
Li et.al~\cite{li2020recurrent} propose a recurrent (\textit{i.e.}~iterative) inpainting method inspired by how humans inpaint from the outer regions towards the inner ones. They progressively reduce the size of the hole by exploiting the correlation between neighboring pixels and strengthen the constraints for estimating deeper pixels.  A Knowledge Consistent Attention module is further utilized that recurrently estimates at each step the attention score for the hole by taking into account the score at the previous step.

As mentioned earlier though, the inpainting task focuses on plausibility and not necessarily the restoration of the real content.
This was the main focus of the work of Zheng et al.~\cite{zheng2019pluralistic} where a probabilistically principled framework was designed to generate multiple plausible results with reasonable content for each masked input.
To achieve that, it combined both generative and variation synthesis approaches.  
Two generators with shared weights were utilized, with the first taking the masked image as input and sampling the encoding vector from a learned probability distribution, which is subsequently decoded to produce the output image. The second one, used only during training, jointly leverages the masked regions and the feature maps of the first decoder to generate the inpainted result.

Taking into consideration the inherent ambiguity of the image inpainting task, generating plausible realistic images based only on reconstruction losses at pixel ($L1$ loss) or at feature level \cite{johnson2016perceptual,gatys2016image} is not feasible. 
In cases where novel contents are missing from the input image, the goal is to generate visual plausible textures, coherent with surrounding known regions while in parallel maintaining the global semantic structure of the image. 
In order to hallucinate such contents, the generative models' contribution is crucial. 
The established zero-sum game between the generator and discriminator enforces the former to synthesize crisp images that cannot be distinguished from the natural image distribution.
Yet this is accomplished in a pure learning framework, with no additional constraints or guarantees of structural alignment.

\textbf{Boundary Preserving Inpainting.}
One important component of photorealism is the preservation of boundaries.
In \cite{li2019progressive} the partial convolution was revisited and used jointly with visual structure reconstruction layers which incorporate structural information in the reconstructed feature map.
This resulted in a progressive joint reconstruction of the visual structure (edges) and features in a progressive manner.
Also, EdgeConnect~\cite{nazeri2019edgeconnect} introduced an edge generator to hallucinate edges in the missing regions which afterward act as structural guidance for the inpainting task.
Likewise, StructureFlow~\cite{Liu2019MEDFE} utilize a two-stage network which consists of a structure reconstructor based on \cite{xu2012structure} and a texture generator. 
The former produces a smooth image with preserved strong edges, while the latter employs appearance flow \cite{zhou2016view} to deliver realistic texture.
Nevertheless, the edge generator in EdgeConnect discards useful information such as image color whilst StructureFlow uses the input image structure only in the first layer, with no guarantees of structural alignment in the deeper layers.

\textbf{Image-to-image Translation.}
Similar to DR inpainting, structure and boundary preservation is very important in the context of image-to-image translation \cite{isola2017image}, a related synthesis task.
Isola et al.~\cite{isola2017image} introduced \textit{pix2pix}, which utilized an image-conditional GAN for multiple applications such as transforming semantic label to  photos, sketches to shoes and Google maps to satellite views. 
One specific application, reconstructing images from semantic labels \cite{chen2017photographic,dong2017semantic,karacan2016learning}, especially focuses on preserving the boundaries between different classes.
As a result, even the earlier attempts \cite{chen2017photographic} reused the boundary highlighting semantic map in multiple stages within their architectures.
Recently, the SPADE blocks \cite{park2019semantic} employed them within a spatially-adaptive normalization layer to propagate the semantic information throughout the network. 
Specifically, the activations in normalization layers are modulated by the semantic segmentation map through a spatially adaptive learned transformation.
Contrary to \textit{pix2pix}, the proposed method does not normalize the input semantic segmentation mask, thus semantic information is better preserved.
Nevertheless, SPADE uses just one style code to control the style of the image without inserting style information throughout the network but only at the beginning. 
To overcome this limitation, SEAN~\cite{zhu2020sean} embeds one style reference for each semantic class and consequently, leverages the style information in the form of spatially-varying normalization parameters.

In our work, we bring the best of both worlds (photorealism and boundary preservation) in a DR-oriented inpainting task.
We exploit the photorealistic generation capacity of generative models in combination with the boundary preserving properties of image-to-image translation task, to replace parts of a \360 image with the occluded background.
By integrating layout estimation, our goal is to synthesize content that is conditioned on it, therefore preserving the actual scene's structure, taking a step away from plausibility towards realism.

\section{Approach}

Our goal is to remove (\textit{i.e.}~diminish) objects in  spherical panoramas of indoor scenes.
On the one hand, the regional nature of our problem is similar to image inpainting, as part of the content needs to be photorealistically hallucinated.
On the other hand, its context necessitates counterfactual predictions, specifically to hallucinate occluded areas, drawing away from image inpainting, towards image-to-image translation, where only specific traits/parts of the original image need to be preserved, and others need to be adapted to another domain.
More specifically, for indoors DR we consider the task of translating a scene filled with objects/foreground to an empty, background only, scene.
At the same time, another requirement imposed by the downstream applications related to interior (re-)planning, is the preservation of \textit{reality}, which, apart from the appearance of the scene, also corresponds to the structure of the scene which needs to be respected when planning changes, as it is largely unchangeable.
To address this particular problem, we follow a hybrid approach that will be described in this section, to photorealistically hallucinate occluded content from masked regions, while respecting the underlying scene structure.
First, in Section~\ref{sec:dataset} we describe the data we use, then in Section~\ref{sec:model} we provide the details of our hybrid structure-preserving counterfactual generative approach for \360 DR, and, finally, in Section~\ref{sec:supervision} we present the model's supervision scheme.

\subsection{\360 Diminished Reality Dataset}
\label{sec:dataset}
\begin{figure*}
    \centering
    \begin{subfigure}{.19\textwidth}
    \includegraphics[width=\textwidth]{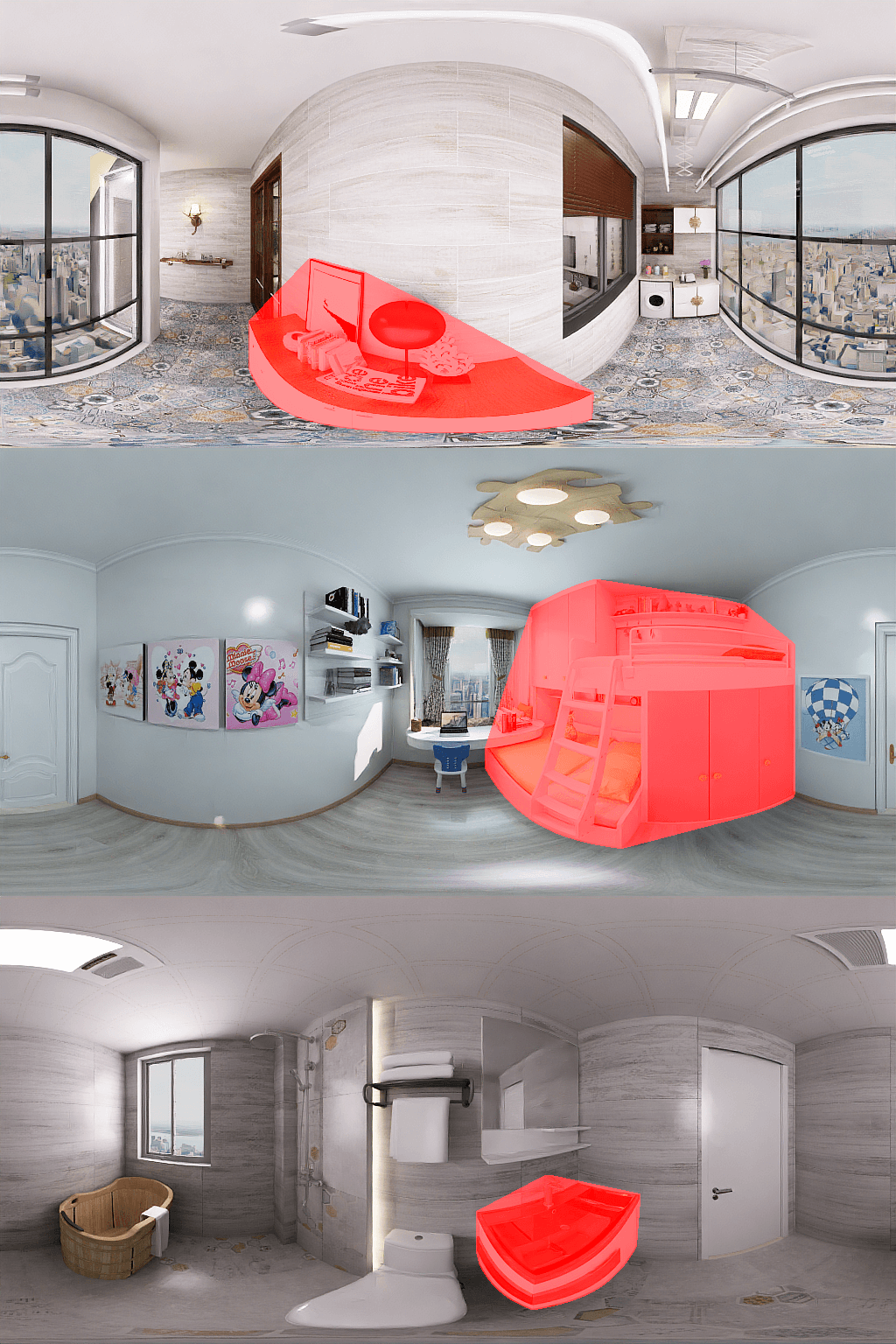}
    \caption{}
    \label{fig:dataset_a}
    \end{subfigure}
    \begin{subfigure}{.19\textwidth}
    \includegraphics[width=\textwidth]{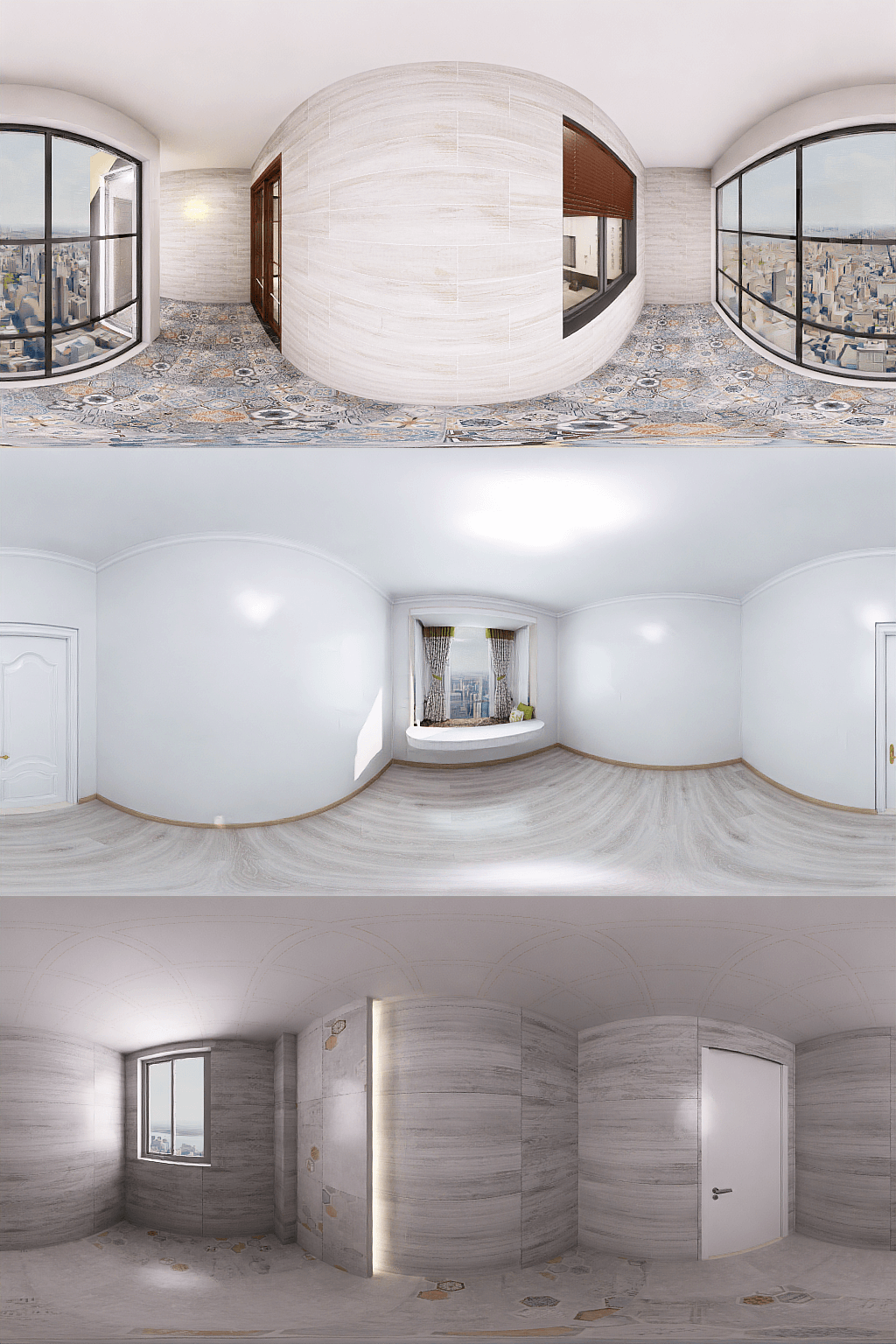}
    \caption{}
    \label{fig:dataset_b}
    \end{subfigure}
    \begin{subfigure}{.19\textwidth}
    \includegraphics[width=\textwidth]{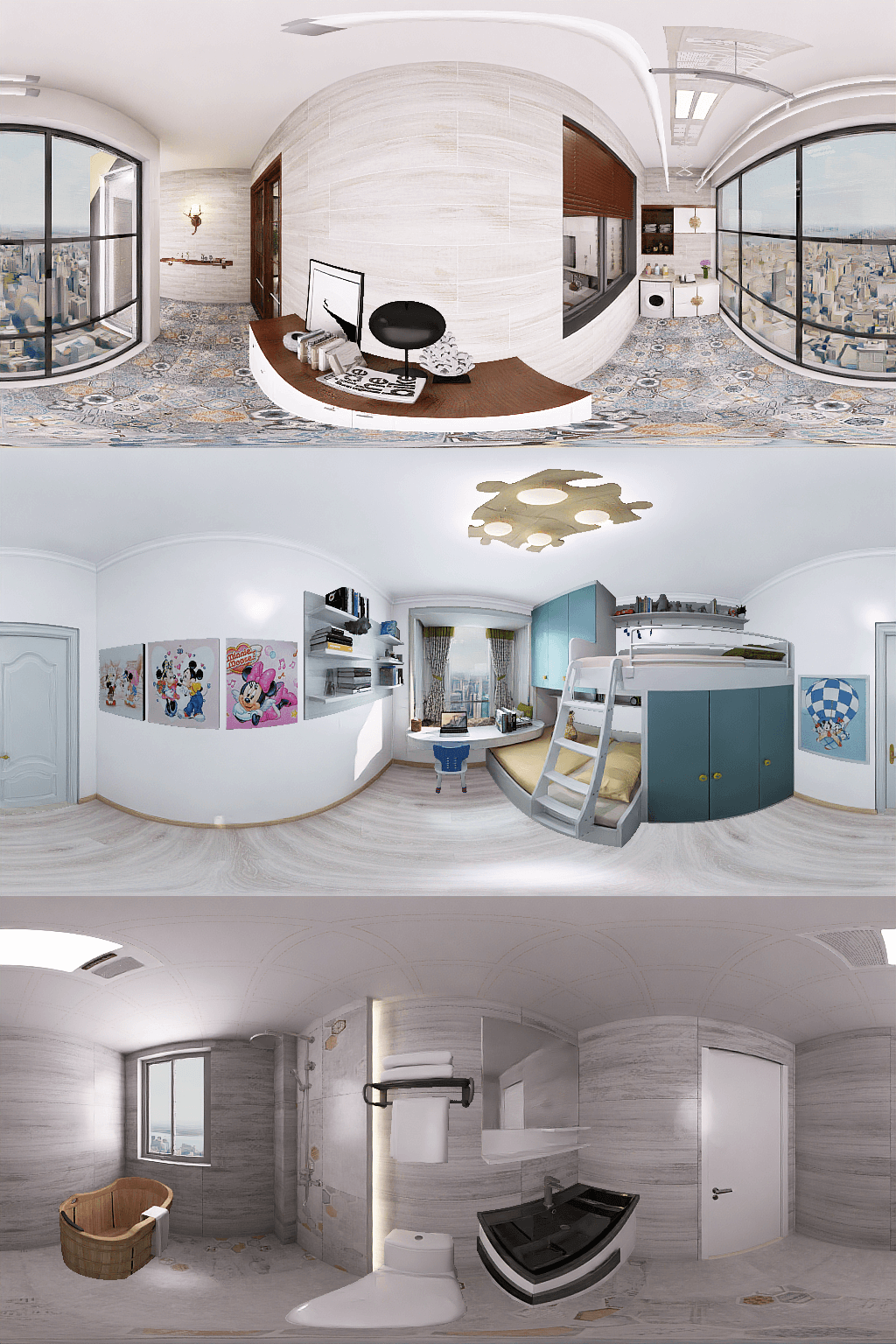}
    \caption{}
    \label{fig:dataset_c}
    \end{subfigure}
    \begin{subfigure}{.19\textwidth}
    \includegraphics[width=\textwidth]{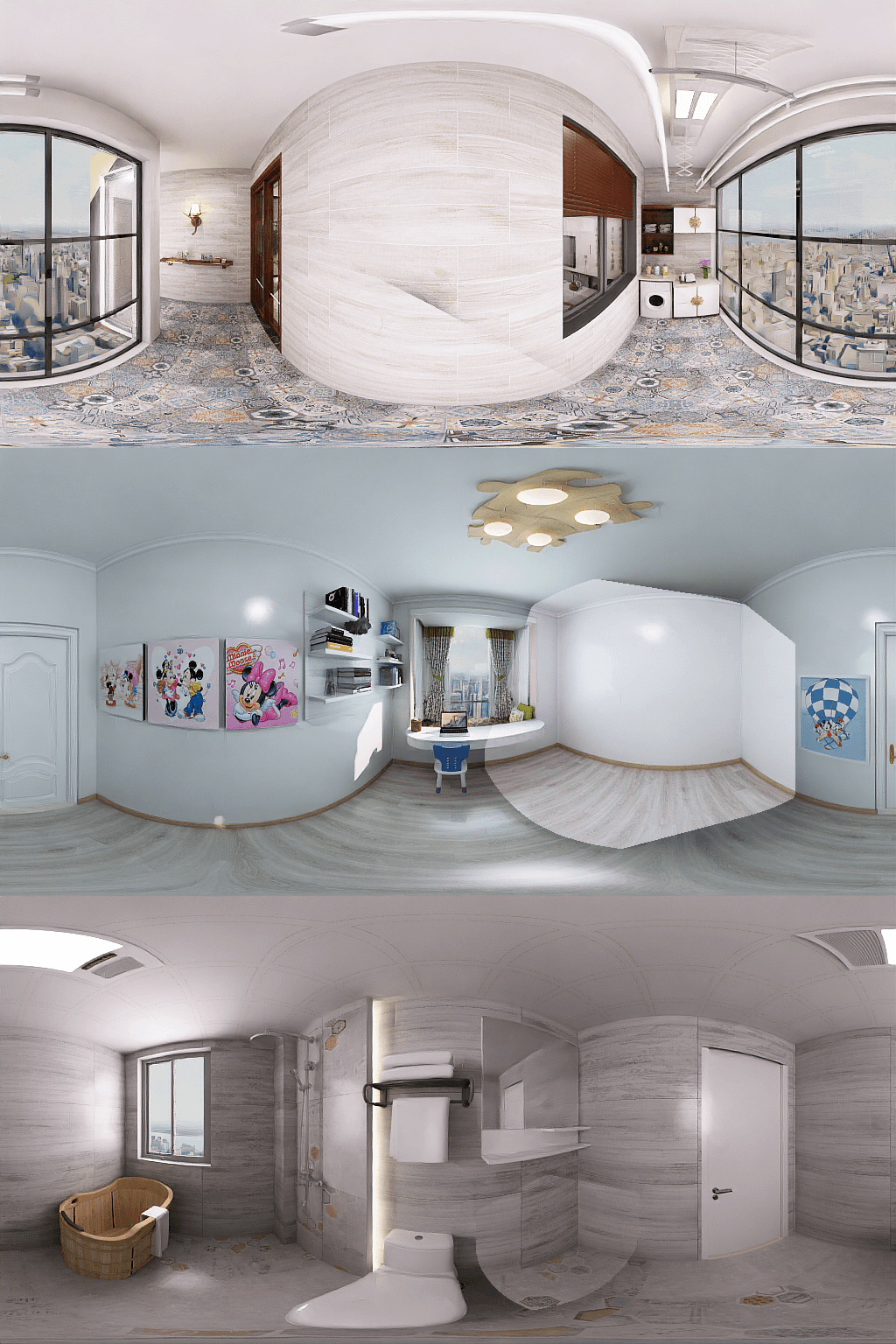}
    \caption{}
    \label{fig:dataset_d}
    \end{subfigure}
    \begin{subfigure}{.19\textwidth}
    \includegraphics[width=\textwidth]{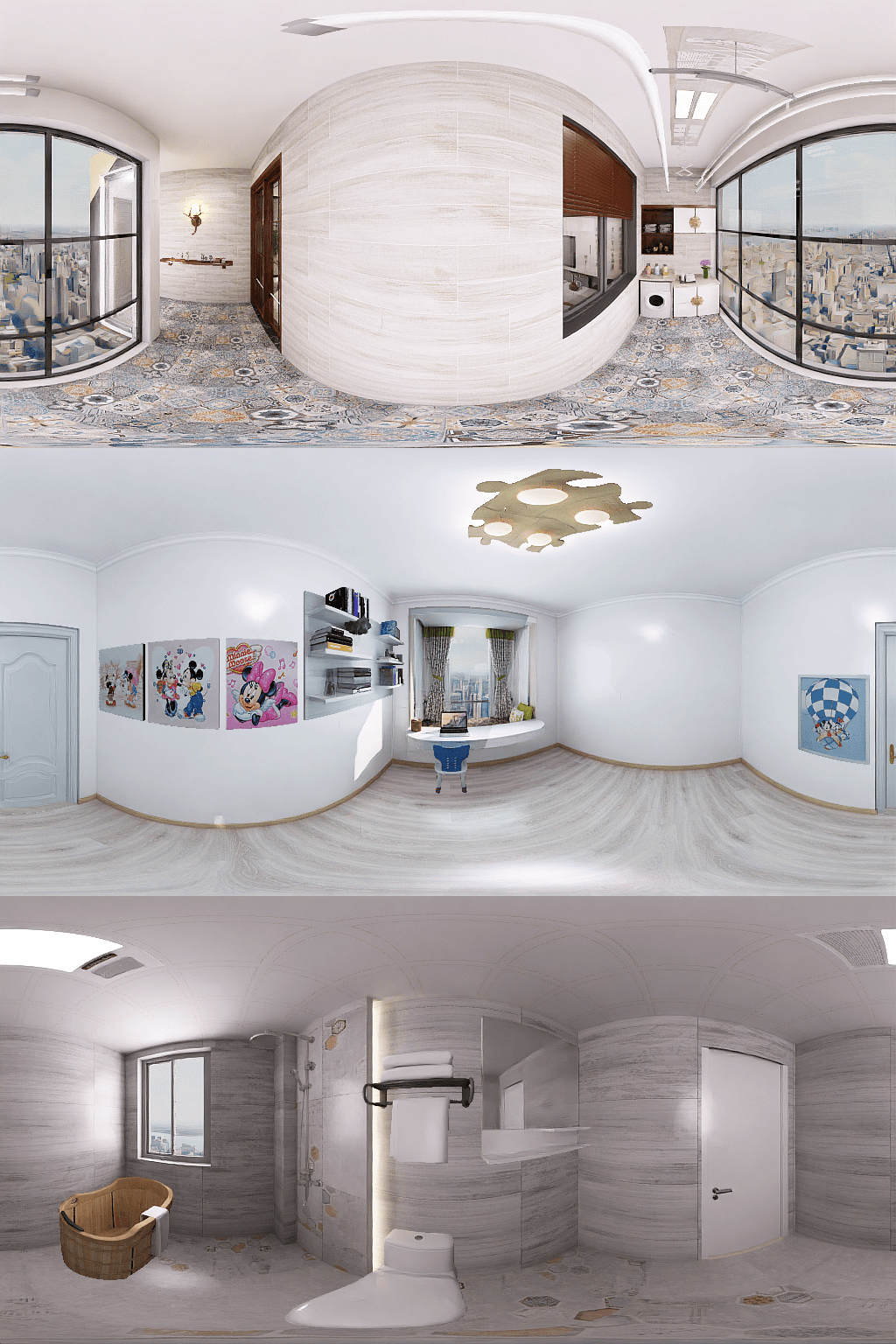}
    \caption{}
    \label{fig:dataset_e}
    \end{subfigure}
    \caption{(a) original sample with highlighted object mask, (b) empty room, (c) full room with empty supervision, (d) augmented room with objects, (e) augmented room with empty supervision}
    \label{fig:dataset}
\end{figure*}

While there exists a variety of datasets of indoor scenes like Matterport3D \cite{chang2018matterport3d} and Stanford2D3D \cite{armeni2017joint}, also including spherical panoramas, they are not suitable for DR.
The task that we tackle in this work is fundamentally different from all the previously explored methods for the simple reason that removing an object from an image must be supervised with the occluded content.

Therefore, we employ the Structured3D dataset \cite{Structured3D}, which provides, among others, photo-realistic spherical panoramas of indoor scenes with 3 different configurations (\textit{empty, simple, and full room}), accompanied by layout and semantic label annotations, making it highly suitable for DR object removal.
Considering a normalized indoor scene color image $\im_f \in \mathbb{R}^{W \times H \times 3} \, \, | \, \, 0 \leq \im_f(\mathbf{p}) \leq 1$, with $\mathbf{p} \in \Omega$, and $\Omega$ being the image domain of width $W$ and height $H = \frac{W}{2}$, which corresponds to either the simple or full configuration of Structured3D, we seek to remove a foreground object by replacing its appearance with the occluded background.
The latter corresponds to the matching scene empty image $\im_b$, which contains a minimal, structure only, WCF (wall--ceiling--floor) representation of the same scene.
In addition, Structured3D offers the scene layout's junction positions, which after projecting the connected wireframe reconstruction on the panorama, provide a dense layout WCF segmentation $\layout \in \mathbb{N}^{W \times H}$ of the $\im_{f/b}$ scene, with $\layout(\mathbf{p}) \in \{ 1, 2, 3 \}$.

To learn the removal of objects, we additionally exploit the semantic labeling of each scene.
We randomly pick the largest connected component of one of the available foreground classes as the input object mask $\obj$ denoting the region to be diminished.
The mask $\obj$ is a binary mask with ones in the diminished region and zeros elsewhere, while $\bar{\obj}$ is its binary inverse mask.
Since the masks are pixel perfect, we calculate their convex to simulate a generic, convex polygon region selection. 
This way, we can supervise the DR task for the diminishing of an input image $\im_f$ at the region denoted by $\obj$, using the background image $\im_b$.
However, this straightforward way fails in practice because of conflicting supervision signals.
Structured3D is photorealistically rendered using physically based ray-tracing.
As a result, the inclusion of foreground introduces light ray bounces, and more importantly, the lights themselves, which are added into the scene as new foreground objects participate into the new image formation process, creating a photo-inconsistency between $\im_f$ and $\im_b$.
To overcome this issue, we perform reverse compositing of selected foreground object classes (\textit{i.e.}~excluding lights) into the background.
While this is not a perfect solution as incident shadows are lost, it allows for proper training that is not hindered by irregular supervision across the training samples.
An illustrative example of this process and the defects it solves is presented in Figure~\ref{fig:dataset}.

\begin{figure*}
    \centering
    \vspace{-0.25cm}
    \includegraphics[width=\textwidth]{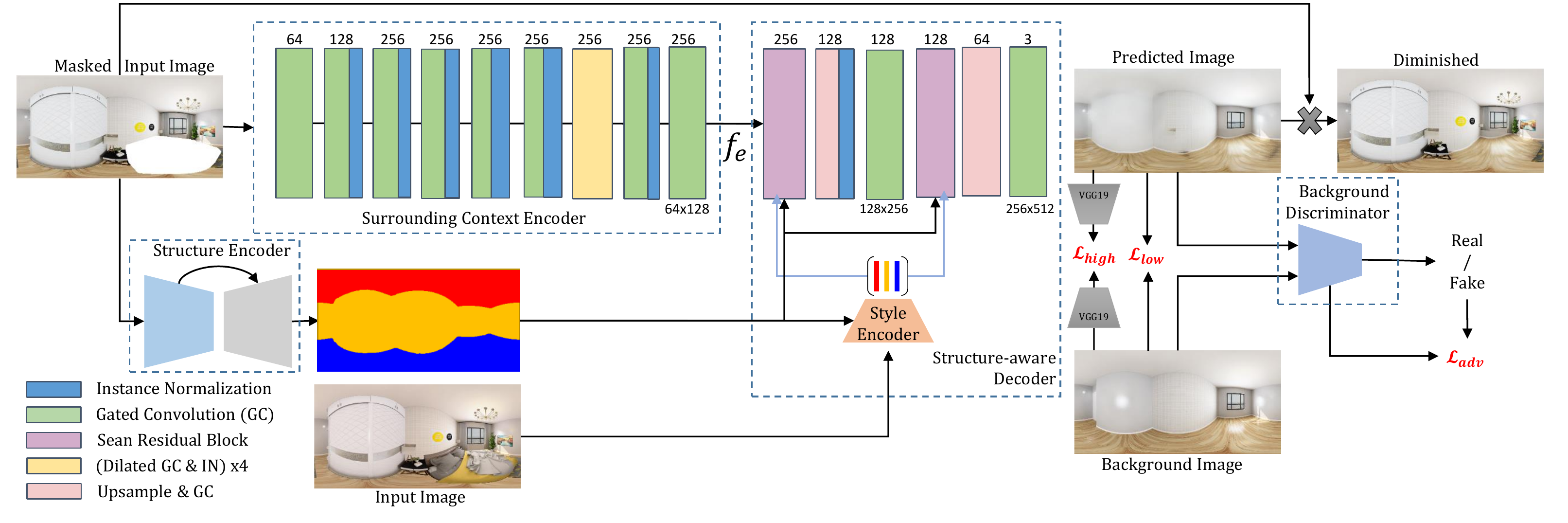}
    \caption{
    The PanoDR model that preserves the structural reality of the scene while counterfactually inpainting it.
    The input masked image is encoded twice, once densely by the structure UNet encoder outputing a layout segmentation map, and once by the surrounding context encoder, capturing the scene's context while taking the mask into account via series gated convolutions.
    These are then combined by the structure-aware decoder with a set of per layout component style codes that are extracted by the complete input image.
    Two SEAN residual blocks ensure the structural alignment of the reconstructed background image that is supervised by low- and high-level losses, as well as an adversarial loss driven by the background image discriminator.
    The final diminished result is created via compositing the predicted and input images using the diminishing mask.
    }
    \label{fig:networks}
    \vspace{-0.25cm}
\end{figure*}

\subsection{Structure-Disentangled DR Model}
\label{sec:model}
The DR task blurs the line between image inpainting/completion and image-to-image translation.
While image completion is among the top trends in the computer vision research, its standard approach is well formulated. 
More precisely, either pre-defined shape masks (\textit{e.g.}~boxes) are used, or free form ones, to corrupt the image, and then supervise learning with the original image.
Apart from the minor mask shape difference, DR predictions are counterfactual as they seek to reconstruct occluded areas, but similar to inpainting, it needs to exploit the context of the scene to diminish it.

Image-to-image translation on the other hand adapts the entire context of an image by translating it to another domain, like photo-to-sketch, or labels-to-image \cite{lu2018image, wang2018high}.
Usually, only part of the context needs to be preserved, most usually the dominant structure.
For our DR case, the masked region needs to be infilled with the occluded context.
While this can be considered as another domain, it is nonetheless closer to the original domain that traditional image-to-image translation tasks, which means that it can be partly inferred from its surrounding context, compared to a translated one.

\textbf{Approach.}
We employ a hybrid approach taking the masked image $\im_m = \im_f \odot \bar{\obj} + \mathbf{1} \odot \obj$ as input, with $\odot$ denoting element-wise multiplication and $\mathbf{1} \in \Omega$ is an all-ones matrix.
Our model generates an output image $\hat{\im_e}$ that with is the empty representation of the original input room.
The final diminished image $\hat{\im_d}$ is then composited as $\hat{\im_d} = \im_f \odot \bar{\obj} + \hat{\im_e} \odot \obj$.

To preserve the structural reality, our approach disentangles the structure in an explicit manner, as well as the style, which gets further disentangled per structural element.
These are then re-entangled when generating the final background image $\hat{\im_e}$, in order to preserve the structure as faithfully as possible, and to generate the appearance of each structural element distinctly.
For image inpainting this is usually done implicitly within the network, even in more recent works \cite{nazeri2019edgeconnect} that seek to respect the content's structure.
For image-to-image translation, boundary information is more important, but the assumption is that the entire image is translated, and that there are no invalid regions like holes that need to be completed.

\textbf{Structure Encoding.}
We explicitly encode the scene's structure by predicting a dense layout segmentation map $\hat{\layout} \in \Omega$ splitting the panorama into $3$ structured regions.
For this we use a well established \textit{UNet} segmentation architecture \cite{ronneberger2015u}, essentially converting structure encoding into a dense classification task. 
The input to the network is $\im_m$, and it consists of 4 down-sampling and up-sampling convolutional modules joined by skip connections, and uses batch normalization and ReLU activations.
The choice of the UNet, with skip connections and $1\!\times\!1$ prediction layers offers finer segmentation results which are very important as they allow for pixel level structure decoupling.

\textbf{Surrounding Context Encoding.}
To encode the surrounding context we use an image inpainting derived architecture, which, specifically, is adapted from \cite{yu2019free}.
Its detailed architecture is depicted in Figure~\ref{fig:networks}, and relies on Gated Convolutions, which are a generalization of partial convolutions \cite{liu2018image} that integrates a learnable gating technique when selecting features.
In addition, instance normalization \cite{ulyanov2016instance} and ReLU activations are used.
Taking into account the spherical nature of our inputs, we circularly pad \cite{sun2019horizonnet} in the horizontal image direction all convolution inputs to overcome the longitudinal boundary discontinuity, and use reflection padding to simulate the singularities at the poles \cite{zioulis2021single}.
For the bottleneck we rely on repeated dilations \cite{yu2017dilated} to capture the global context more efficiently by expanding the receptive field, avoiding additional  parameters  and  preventing  immoderate  information  loss.
This inpainting-derived encoder extracts the content excluding the hole and outputs features $\mathbf{f}_e$.

\textbf{Structure-aware Decoding.}
The decoder uses a cascade of SEAN residual blocks \cite{zhu2020sean}, gated convolutions and upsample layers.
The inputs to the decoder are the predicted structure map $\hat{\layout}$, the global context features $\mathbf{f}_e$ as encoded by the surrounding context encoder, and the original -- unmasked -- image $\im_f$.
The latter offers a set of style codes $\mathbf{f}_s^i$ for each structural element $i \in \{wall, ceiling, floor\}$ as learned by a shallow ``bottleneck'' convolutional autoencoder with a label-wise average pooling output layer.
Compared to the global context $\mathbf{f}_e$, these style codes encode local features, corresponding to texture-like details.
Through the use of SEAN blocks, we re-entangle the dense global structure $\hat{\layout}$, the global surrounding context $\mathbf{f}_e$, and the local style of each structural element $\mathbf{f}_s^i$.
This way, we condition our decoder to respect the global structure while completing the translated image, and modulate the inpainted region's style with the style codes extracted for each structural element.

\textbf{Background Discriminator.}
We use a discriminator to adaptively learn the differences between the translated empty images and the corresponding ground truth ones.
More specifically, we use a global PatchGAN \cite{isola2017image} discriminator with spectral normalization \cite{miyato2018spectral}. 
Its input is either the decoded output $\hat{\im_e}$ or the empty background image $\im_e$ concatenated with the mask $\obj$ and classifies each patch of the input image as real or fake. 
Its output $\mathbf{d}$ is a score map rather than a single score, where each value corresponds to a local region of the input sample covered by its receptive field.

\subsection{Supervision}
\label{sec:supervision}
The structure encoding model is supervised with binary cross-entropy, as it is formulated as a dense classification task.
For the final inferred background image we use a combination of different domain losses to ensure the photorealistic quality of the predictions:
\vspace{-0.15cm}
\begin{equation}
    \loss = \loss_{low} + \loss_{high} + \loss_{adv}.
    \vspace{-0.15cm}
\end{equation}
A low level reconstruction loss $\loss_{low}$, a high level synthesis loss $\loss_{high}$, and an adaptive adversarial loss $\loss_{adv}$.

\textbf{Low-level Reconstruction Loss.}
This pixel-based loss focuses on the reconstruction of low frequency components of the predicted image $\hat{\im_e}$:
\vspace{-0.15cm}
\begin{equation}
\label{eq:reconstruction_loss}
    \loss_{low} = \lambda_{L1} \frac{1}{N} \sum_{\mathbf{p}}^{\Omega} \Attention | \im_{e} - \hat{\im_{e}} | + \lambda_{tv} \obj (| \nabla_x \hat{\im_e} | + | \nabla_y \hat{\im_e} |),
\end{equation}
where $N$ is the total number of pixels, and $\Attention(\mathbf{p}) \in \mathbb{R}^{W \times H}$ is the spherical attention mask used in \cite{zioulis2019spherical} that accounts for equirectangular distortion.
Apart from the spherically weighted L1 loss, a total variation smoothness prior is used for the diminished area specifically to counter the high frequency artifacts usually seen in the early training stages of generative models.

\textbf{High-level Synthesis Loss.}
Apart from encouraging $\hat{\im_e}$ and $\im_{e}$ to have the same representation at the pixel level with $\loss_{low}$, we additionally employ a data-driven loss $\loss_{high}$.
This enforces them to have a similar representation in the feature space as computed by a CNN model $\vgg$, which in our case, is a pre-trained VGG-19 \cite{simonyan2014very}. 
Let $\vgg_{j}(\im)$ be the activations of the $jth$ layer of the network $\vgg$, for the given image $\im$, $\Omega_j$ its feature element domain, and $N_j$ the total number of feature elements of the $j$ feature map.
Then the loss is formulated as a combination of the perceptual and style losses:
\begin{align}
\label{eq:synthesis_loss}
    \loss_{high} &= \lambda_{perc} \loss_{perc} + \lambda_{style} \loss_{style}\\
    \loss_{perc} &= \sum_j^{P_j} \frac{1}{N_{j}} \sum_{\boldsymbol{\rho}}^{\Omega_j} | \vgg_{j}(\im_{e}) - \hat{\vgg_{j}(\im_{e})} |\\ \loss_{style} &=\sum_j^{S_j} \frac{1}{N_{j}} \sum_{\boldsymbol{\rho}}^{\Omega_j} \frac{1}{N_{j}} | \pazocal{G}(\vgg_{j}(\im_{e})) - \pazocal{G}(\vgg_{j}(\hat{\im_{e})}) |,
\end{align}
where $P_j$ and $S_j$ are the set of features used for the perceptual and style \cite{gatys2016image, johnson2016perceptual} losses, and $\pazocal{G}(\mathbf{M}) = \mathbf{M}\mathbf{M}^T$ is the Gram matrix function.
Both losses are derived in a high dimensional data-driven feature space, with the former (perceptual) operating on a global level, and the latter (style) operating on global and local levels.

\textbf{Adaptive Adversarial Loss.}
To adaptively improve the quality of the generated background images $\hat{\im_e}$ we additionally employ a discriminator-based loss that is learned during training.
Since we use a PatchGAN disciminator, we formulate our combined adversarial loss as a combination of a hinge loss on the final real/fake predictions \cite{lim2017geometric}, and a feature matching loss using the discriminator's intermediate features:
\begin{align}
\label{eq:adversarial_loss}
    \loss_{adv} &= \lambda_{D} \loss_{D} + \lambda_{FM} \loss_{FM} \\
    \loss_{D} &= \frac{1}{N_d} \sum_{\mathbf{p}}^{\Omega_d} r(1 - \mathbf{d}_e) + \frac{1}{N_d} \sum_{\mathbf{p}}^{\Omega_d} r(1 + \mathbf{d}_{\hat{e}})\\
    \loss_{FM} &= \sum_i^{D_i} \frac{1}{N_d^i} \sum_{\mathbf{p}}^{\Omega_d^i} | \mathbf{d}_e^i - \mathbf{d}_{\hat{e}}^i|,
\end{align}
where $\mathbf{d}_e$ and $\mathbf{d}_{\hat{e}}$ are the discriminator outputs for the real and predicted background images, $\Omega_d$ is the pixel domain of the discriminator's output, $N_d$ the total count of its spatial elements, and the $i$ denotes intermediate discriminator feature maps. 
The spatial discriminator hinge loss and the feature matching loss are weighted by their respective weights.
Feature matching enforces the generator to minimize the statistical difference between the features of the ground truth images and the generated images, which helps further stabilize the training and improve the quality of the generated content.

\section{Results}
\begin{figure*}[!htbp]
    \centering

\subfloat{\includegraphics[width=0.24\linewidth]{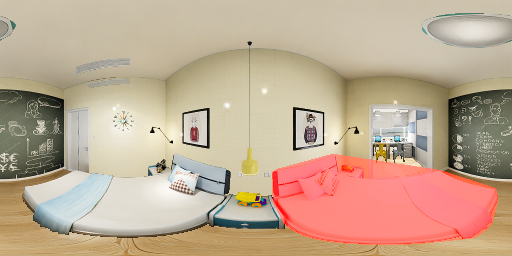}} %
\hfill
\subfloat{\includegraphics[width=0.24\linewidth]{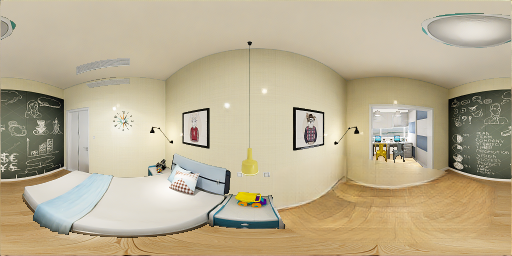}} 
\hfill
\subfloat{\includegraphics[width=0.24\linewidth]{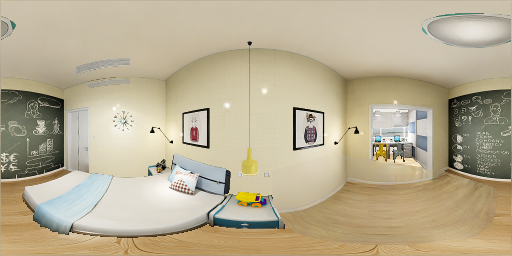}} %
\hfill
\subfloat{\includegraphics[width=0.24\linewidth]{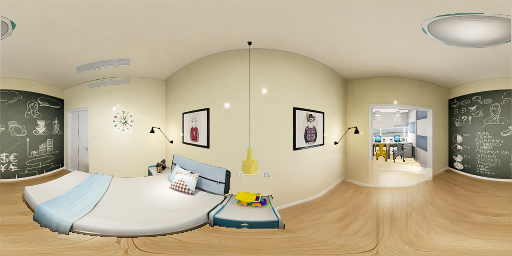}} %
\hfill

\subfloat{\includegraphics[width=0.24\linewidth]{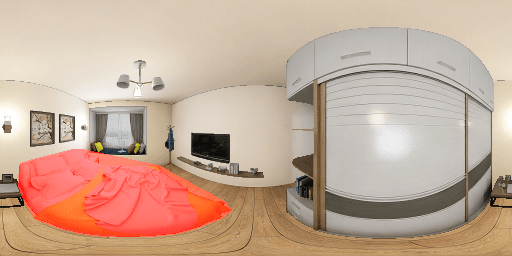}} %
\hfill
\subfloat{\includegraphics[width=0.24\linewidth]{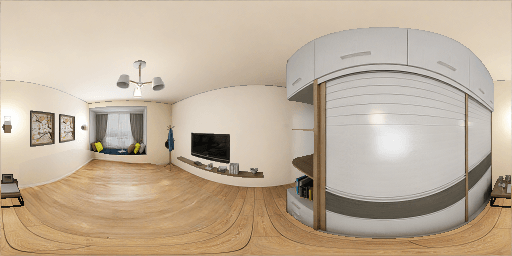}} %
\hfill
\subfloat{\includegraphics[width=0.24\linewidth]{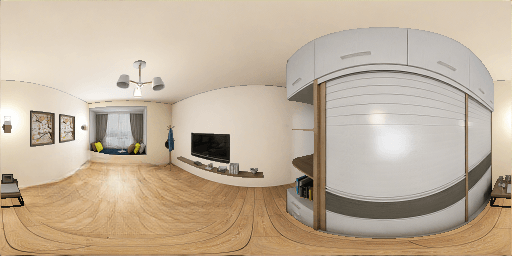}} %
\hfill
\subfloat{\includegraphics[width=0.24\linewidth]{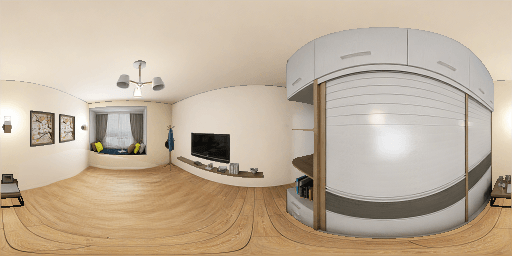}} %
\hfill
\subfloat{\includegraphics[width=0.24\linewidth]{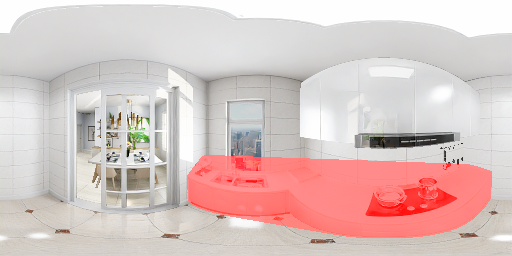}} %
\hfill
\subfloat{\includegraphics[width=0.24\linewidth]{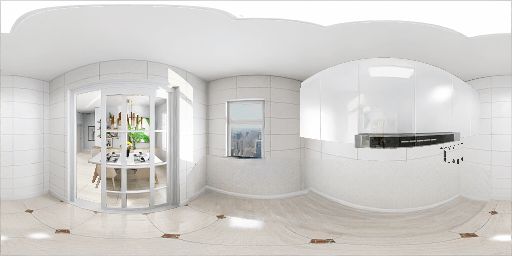}} %
\hfill
\subfloat{\includegraphics[width=0.24\linewidth]{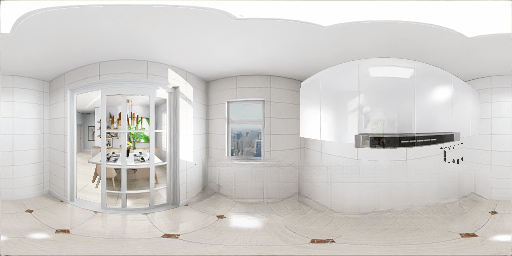}} 
\hfill
\subfloat{\includegraphics[width=0.24\linewidth]{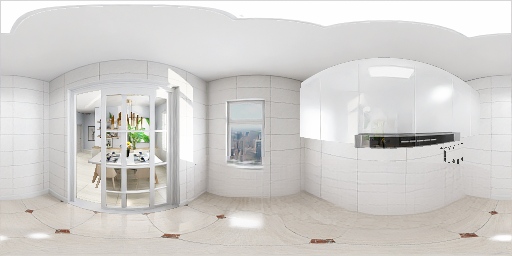}} %
\hfill

\subfloat{\includegraphics[width=0.24\linewidth]{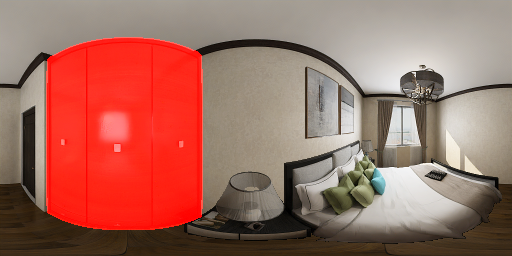}} %
\hfill
\subfloat{\includegraphics[width=0.24\linewidth]{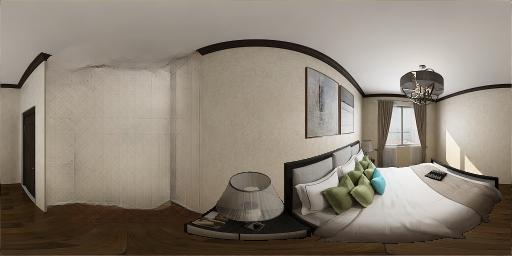}} %
\hfill
\subfloat{\includegraphics[width=0.24\linewidth]{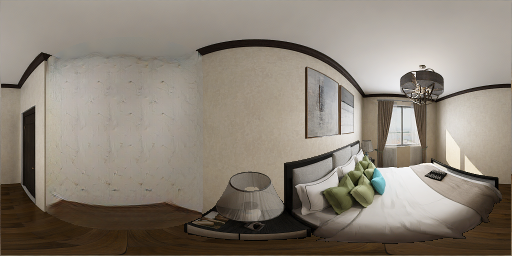}} 
\hfill
\subfloat{\includegraphics[width=0.24\linewidth]{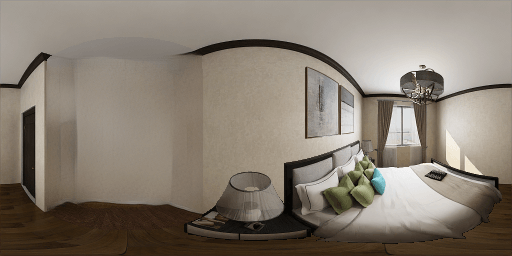}} %
\hfill
\subfloat{\includegraphics[width=0.24\linewidth]{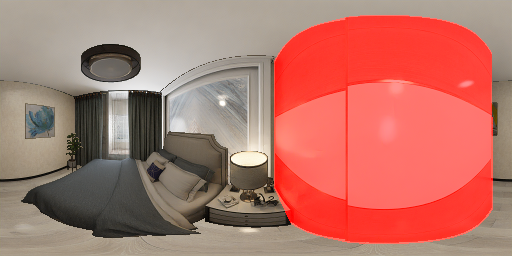}} %
\hfill
\subfloat{\includegraphics[width=0.24\linewidth]{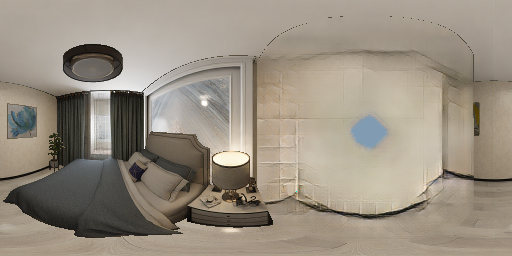}} %
\hfill
\subfloat{\includegraphics[width=0.24\linewidth]{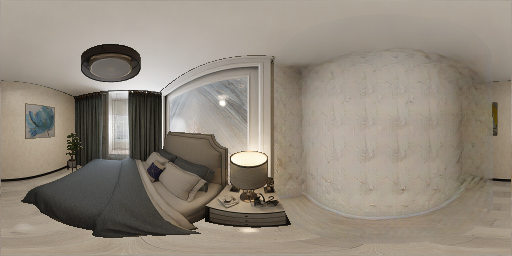}} 
\hfill
\subfloat{\includegraphics[width=0.24\linewidth]{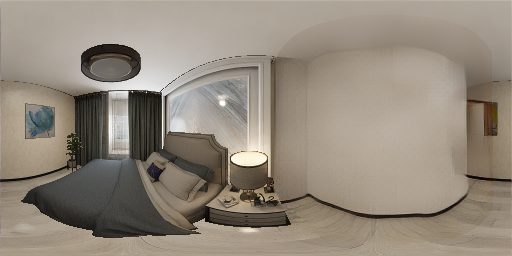}} %
\hfill

\subfloat{\includegraphics[width=0.24\linewidth]{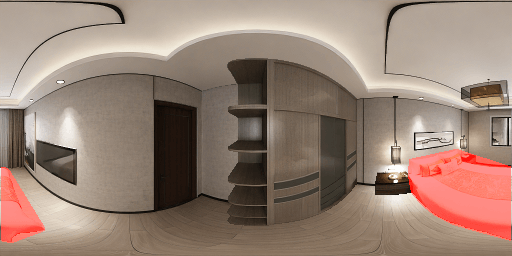}} %
\hfill
\subfloat{\includegraphics[width=0.24\linewidth]{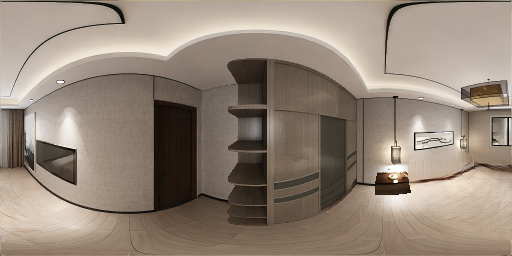}}
\hfill
\subfloat{\includegraphics[width=0.24\linewidth]{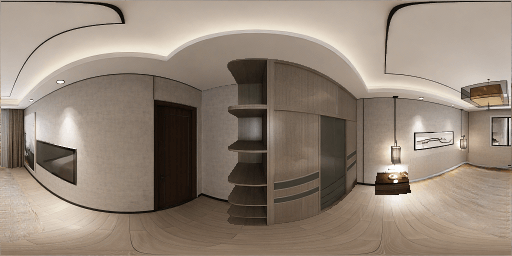}} %
\hfill
\subfloat{\includegraphics[width=0.24\linewidth]{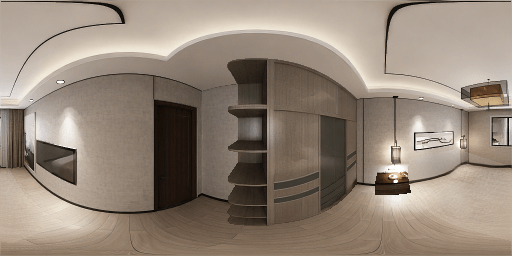}} %
\hfill
\vspace{-0.35cm}
\caption{
Qualitative results for diminishing objects from scenes in our test set.
From left to right: Input image with the diminished area masked with transparent red, RFR, PICNet and ours. 
}
\label{fig:qualitative}
\vspace{-0.33cm}
\end{figure*}

\textbf{Implementation Details.}
We implement our model using Pytorch \cite{paszke2017automatic} with all experiments conducted on a Nvidia GeForce RTX 3090 GPU.
Our generative models are optimized using Adam \cite{kingma2014adam}, with $b_{1}=0.5$ and $b_{2}=0.999$, a learning rate of $0.0002$ and a batch size of $6$. 
The segmentation UNet is optimized with a default parameterized Adam using a learning rate of $0.0001$ and a batch size of $4$.
The input and output panorama resolutions are $512 \times 256$.
The weights of the UNet are initialized with \cite{he2015delving} and for the other sub-models from a zero-centered Normal distribution with $\sigma=0.02$. 
We empirically set $\lambda_{L1} = 4.0$, $\lambda_{TV} = 1.0$, $\lambda_{perc} = 0.15$, $\lambda_{style} = 40.0$, $\lambda_{D} = 0.2$ and $\lambda_{FM} = 20.0$.

\begin{table}[!htbp]
\centering
\caption{
Quantitative results assessing photorealism (LPIPS, PSNR, SSIM, MAE) and structural preservation (mIoU) on the S3D test set. 
}

\label{tab:quantitative_results}
\resizebox{\linewidth}{!}{%
\begin{tabular}{@{}l|c|c|c|c | c | c}
\toprule
\textbf{Method}   &  \textbf{LPIPS $\downarrow$}  & \textbf{PSNR $\uparrow$} & \textbf{SSIM $\uparrow$} & \textbf{MAE $\downarrow$} & \textbf{mIoU $\uparrow$} \\ \midrule
RFR \cite{li2020recurrent} & 0.0510 & 31.0114 & 0.9528 & 0.0067  & 0.8583     \\ \midrule
PICNet \cite{zheng2019pluralistic}   & 0.0533 & 32.3072 & 0.9557 & 0.0070  & 0.8502\\ \midrule
Ours  &  \textbf{0.0398} & \textbf{33.6611}  & \textbf{0.9620}  &\textbf{0.0058}  & \textbf{0.8768} \\ \midrule
\end{tabular}
}
\end{table}

\begin{figure*}[!htbp]
    \centering    
    \begin{subfigure}{.19\textwidth}
    \includegraphics[width=\textwidth]{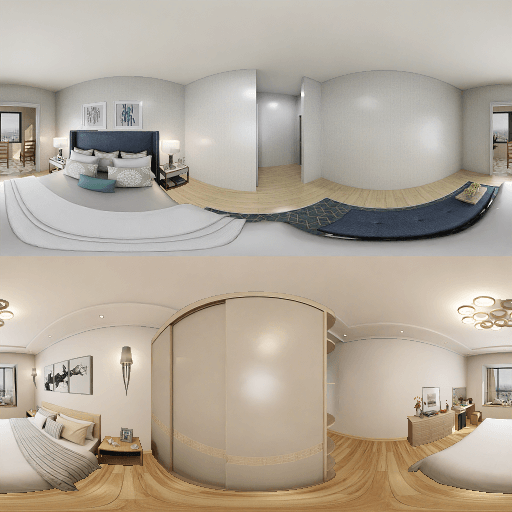}
    \subcaption{}
    \end{subfigure}
    \begin{subfigure}{.19\textwidth}
    \includegraphics[width=\textwidth]{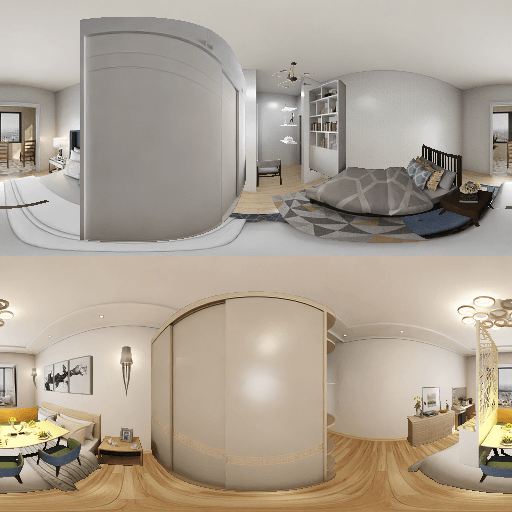}
    \subcaption{}
    \end{subfigure}
    \begin{subfigure}{.19\textwidth}
    \includegraphics[width=\textwidth]{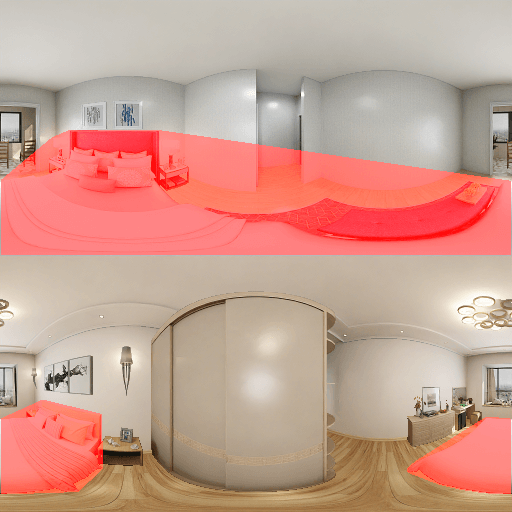}
    \subcaption{}
    \end{subfigure}
    \begin{subfigure}{.19\textwidth}
    \includegraphics[width=\textwidth]{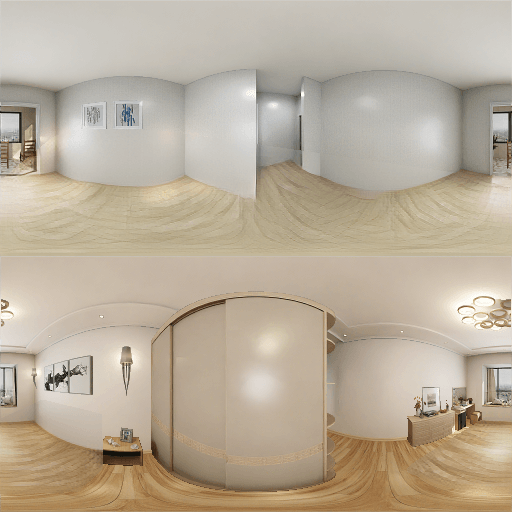}
    \subcaption{}
    \end{subfigure}
    \begin{subfigure}{.19\textwidth}
    \includegraphics[width=\textwidth]{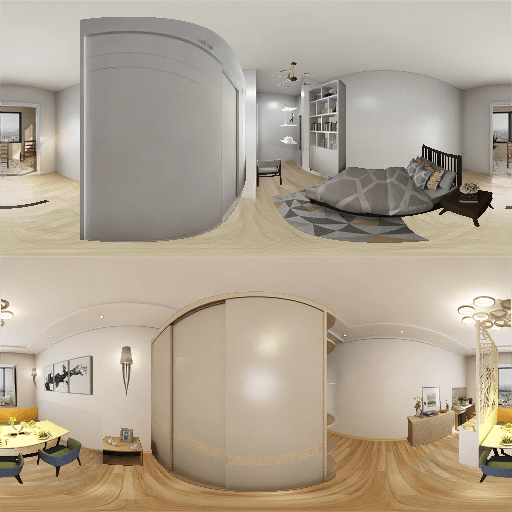}
    \subcaption{}
    \end{subfigure}
    \vspace{-0.35cm}
    \caption{A demonstration of our method, which levitates AR/DR applications. 
    (a) original panorama (b) augmented reality without diminished reality (c) highlighted object for removal (d) our inpainting method result (e) augmented reality with diminished reality, the result is much more natural.}
    \label{fig:application}
    \vspace{-0.35cm}
\end{figure*}

\textbf{Experiments.}
We compare our approach with the following state-of-the-art inpainting methods, PICNet \cite{zheng2019pluralistic} and RFR \cite{li2020recurrent}.
We use their official implementations to train both models on our adapted S3D dataset till convergence using the same empty groundtruth images. 
For our test set we use 569 images containing objects from the official S3D split, with the diminished regions masked as described in Section~\ref{sec:dataset}. 

\textbf{Quantitative Comparison.}
Table~\ref{tab:quantitative_results} compares the photorealistic performance of our model compared to PICNet and RFR.
Standard metrics are used, the Mean Absolute Error (MAE), the Peak Signal-to-Noise Ratio (PSNR), the Structural Similarity Index (SSIM) \cite{wang2004image}, and the Learned Perceptual Image Patch Similarity (LPIPS) \cite{zhang2018unreasonable}.
LPIPS is a metric that has been shown to better assess the perceptual similarity between two images. 
It measures the distance between the target and generated images using features extracted from a pre-trained VGG-16 model. 
Given the generative nature of our task which aims at natural diminishing/counterfactual inpainting, thus generating plausible and photo-realistic content, LPIPS is considered as an important metric for our evaluation.
Nonetheless, our goal is to preserve reality as well, therefore full reference objective measures are also important indicators.
We observe that our approach generates content that is perceptually closer to the natural image distribution as shown by the LPIPS metric and the close to $25\%$ performance gain compared to both PICNet and RFR.
Furthermore, our method is likely to generate smoother results, which are reflected through the per-pixel and pixel neighborhood based objective metrics.

Moreover, in the DR context, the structural boundary preservation is also of interest.
To assess performance with respect to that, we train a layout segmentation model without holes, but with the same training setup as the structure encoder (Section~\ref{sec:model}), and use it to measure the mean intersection-over-union (mIoU) between the groundtruth and the diminished/inpainted results from each respective model.
The evaluation is focused on the segmentation results inside the mask $\obj$.
As indicated in Table~\ref{tab:quantitative_results}, our model preserves the structural boundary more consistently than PICNet and RFR alike.

\textbf{Convergence Analysis.}
It should also be noted that our approach exhibits significantly faster convergence.
The results for PICNet and RFR are trained for $160$ and $190$ epochs respectively, while our model is only trained for $60$ epochs.
This is attributed to the interaction between the structure preserving SEAN blocks and the discriminator.
Longer trains are directly related to the discriminator's performance and added benefits.
The adaptive nature of a learnable adversarial loss helps continuously improve results.
This adaptation is phased, first focusing on coarse structure, and progressively adapting to finer details as the generative models learns to consistently output coherent structures.
However, with our explicit structure reasoning and integration in the decoder, our adversarial loss -- supported by the other low and high level losses -- can quickly start discriminating details, allowing the model to converge faster to high quality results.

\textbf{Qualitative Comparison.}
Figure~\ref{fig:qualitative} presents a set of qualitative results depicting the hallucinated content from our method compared to PICNet and RFR.
Our result exhibit photo-realistic textures and structures which are coherent with the background of the images.
More specifically, the composite photorealism loss allows our method to generates plausible textures. 
Further, it is evident in both Figure~\ref{fig:teaser} (where it is highlighted) and Figure~\ref{fig:qualitative}, that the explicit SEAN blocks guided by the layout segmentation results better preserve each scene's structural boundaries when counterfactually completing it. 
In contrast, the compared methods exhibit some flaws, especially at surfaces' boundaries due to the absence of structural guidance.
Unsurprisingly, in cases where compared methods generate such artifacts, our method can synthesize realistic textures with plausible structures, driven by the disentangling of the structure, and each structural components style as provided by the structure and style encoders and the SEAN blocks.

\textbf{DR-enhanced AR.}
Indoor DR technology is important for supporting AR planning applications.
While AR can support the addition of new objects into the scene, it fails to achieve its goal of enhancing the planning experience when seeking to replace objects.
For such cases, DR can first be applied on the content that is to be replaced, and then AR can composite the new virtual objects.
Figure~\ref{fig:application} shows this specific use case where DR technology is highly relevant.
For these indoor planning use cases, preserving the scene's structure is very important as it allows for productive interactivity without losing the important context.

\section{Conclusion}
Diminishing reality is a very challenging task as the hallucination of reality in counterfactual settings is hard to constrain.
We show that when considering reality in specific contexts, like indoor planning, sufficient preservation of reality is possible.
We combine recent advances in two synthesis tasks and a novel dataset to demonstrate increased performance for spherical panorama based DR.
Our approach relies on the segmentation results, which is a limitation given that the fragility of one sub-model is heavily inter-twinned with the resulting diminishing performance.
One line of research forward for this task would be complete full-to-empty image translation, which would, nonetheless, require an end-to-end model that would be able to separate the foreground from the background.

\textbf{Acknowledgements.}
This work was supported by the EC funded H2020 project ATLANTIS [GA 951900].

{\small
\bibliographystyle{ieee_fullname}
\bibliography{egbib}
}

\end{document}